\documentclass[letterpaper]{article} 
\usepackage[]{aaai24}  
\usepackage{times}  
\usepackage{helvet}  
\usepackage{courier}  
\usepackage[hyphens]{url}  
\usepackage{graphicx} 
\usepackage{natbib}  
\usepackage{caption} 
\usepackage{algorithm}
\usepackage{algorithmic}
\usepackage{multirow}

\usepackage{newfloat}
\usepackage{listings}
\DeclareCaptionStyle{ruled}{labelfont=normalfont,labelsep=colon,strut=off} 
\floatstyle{ruled}
\newfloat{listing}{tb}{lst}{}
\floatname{listing}{Listing}
\newcommand{\primarykeywords}{
None
}

\usepackage{amsmath}
\title{Large Language Models as Planning Domain Generators}
\author{
    James Oswald\textsuperscript{\rm 1},
Kavitha Srinivas\textsuperscript{\rm 2},
Harsha Kokel\textsuperscript{\rm 2},
Junkyu Lee\textsuperscript{\rm 2},
Michael Katz\textsuperscript{\rm 2},
Shirin Sohrabi\textsuperscript{\rm 2}
}
\affiliations{
    \textsuperscript{\rm 1}Rensselaer Polytechnic Institute\\
    \textsuperscript{\rm 2}IBM Research\\
    oswalj@rpi.edu, \{kavitha.srinivas, harsha.kokel, junkyu.lee, michael.katz1\}@ibm.com, ssohrab@us.ibm.com
}

\begin{document}

\maketitle


\begin{abstract}

Developing domain models is one of the few remaining places that require manual human labor in AI planning. Thus, in order to make planning more accessible, it is desirable to automate the process of domain model generation. To this end, we investigate if large language models (LLMs) can be used to generate planning domain models from simple textual descriptions. Specifically, we introduce a framework for automated evaluation of LLM-generated domains by comparing the sets of plans for domain instances. Finally, we perform an empirical analysis of 7 large language models, including coding and chat models across 9 different planning domains, and under three classes of natural language domain descriptions. Our results indicate that LLMs, particularly those with high parameter counts, exhibit a moderate level of proficiency in generating correct planning domains from natural language descriptions.  
Our code is available at \url{https://github.com/IBM/NL2PDDL}.
\end{abstract}

\section{Introduction}

Large language models (LLMs) have demonstrated robust emergent abilities for open-ended tasks like story generation, poetry, and dialogue~\cite{ZhaoSDMCVBKSF23,ImitationGames}. Their potential is no longer limited to natural language. Rather, they have shown the ability to generate highly structured output that resembles code from natural language descriptions of programs~\cite{Li2023StarCoderM, Touvron2023LLaMAOA}. 
It is natural to wonder how these abilities generalize to knowledge engineering tasks such as those used for problem representation in symbolic methods. 
Despite the efficacy of symbolic methods such as boolean satisfiability (SAT) solvers~\cite{SATBook}, automated planners~\cite{Helmert06}, and automated theorem provers~\cite{HarrisonUW14} in their respective domains, the issue of representing a problem accurately and efficiently still hinders the wider adoption and accessibility of these powerful methods. If LLMs can bridge the gap between natural language description of the problem and symbolic representation, it would enable large-scale adoption of symbolic methods and reduce the dependency on technical experts. 
%
It is natural to look to the emergent abilities of LLMs as a potential bridge that could tie natural language descriptions of problems to accurate symbolic representations for automated planning \cite{GhallabNT}. If LLMs are sufficiently equipped for this task, this would reduce the dependency on technical experts as well as lead to a wider adoption of symbolic methods. To this end, our work investigates using LLMs for generating problem representations for automated planning. Specifically, we evaluate the capabilities of LLMs to automatically translate natural language descriptions of planning domains to Planning Domain Description Language (PDDL) \cite{mcdermott-et-al-tr1998}.

The problem of domain generation from natural language has been studied earlier~ \cite{Lindsay2017,Hayton2020} and recently \citet{Guan2023} also attempted this problem using LLMs. Despite these studies, the task of evaluating the usefulness of the generated domain description is extremely difficult. Previous works leveraged human experts for evaluation. We argue that for rigorous, automated evaluation we need a ground truth; a vetted domain specification. Hence, in this work we focus on the task of creating \emph{high-quality reconstructions} of the PDDL domain from natural language; where the generated domain is ideally equivalent to the ground truth. Restricting the generation of the PDDL domain to an approximated equivalence class would make the generated domains more amenable to existing planners and further the goal of using the generated descriptions for producing executable plans. To further clarify, while \citet{Guan2023} uses LLMs to learn a PDDL model from a textual description, this is not our main purpose in this work. We aim in this work to understand how such methods can be evaluated, and due to this, need to depend on the additional assumption that a reference domain is available. While this is a stronger assumption than what is made in earlier work, this allows for fully automated evaluation.  

The core contributions of this work are fourfold.
First, we define a task of {PDDL domain reconstruction} from natural language; based on a ground truth. 
Second, we define two metrics for evaluating domain quality that do not depend on any form of manual human evaluation. 
Third, we examine classes of natural language descriptions of PDDL actions to investigate if the inclusion and exclusion of particular information impacts the ability to generate domains or the quality of generated domains. 
%
Finally, we evaluate $7$ different LLMs, including coding and chat models,
and provide a detailed analysis of the results from each on $9$ domains.

\section{Background}


\subsubsection{Planning}

In this work, we use the Planning Domain Definition Language (PDDL) for the declarative plan representation, but when necessary to discuss the underlying formalisms we refer to parts of planning problems and domains using the following lifted STRIPS formalism, largely in line with \citet{Corra2022BestFirstWS}. A lifted STRIPS planning problem is defined as a 5-tuple $\Pi = \langle \mathcal{F}, \mathcal{C}, \mathcal{A}, s_0, S_{\ast} \rangle$. $\mathcal{F}$ is a finite set of predicates that describe the world. $\mathcal{C}$ is a finite set of constants representing objects in the world, optionally including type information. We define $\mathcal{F}_g$ as the set of all \textit{grounded predicates}, that is, predicates in which all variables are replaced by legal constants from $\mathcal{C}$. A state $s \subseteq \mathcal{F}_g$ is a set of grounded predicates that describe the state of the world, such that $f \in s$ if and only if $f$ is a true fact about the world. The set of all possible states is the power set of $\mathcal{F}_g$, denoted by $S$.
$\mathcal{A}$ is a set of action schema where each $a \in \mathcal{A}$ is a 3-tuple $\langle pre(a), add(a), del(a) \rangle$ where $pre(a) \subseteq \mathcal{F}$ is the set of predicates that must hold to apply the action, $add(a) \subseteq \mathcal{F}$ is the set of predicates that become true after the action is applied, and  $del(a) \subseteq \mathcal{F}$ is the set of predicates that become false. An action schema $a \in A$ can be grounded by substituting all variables in $a$ with allowed constants from $\mathcal{C}$. The grounded action $a_g = \langle pre_g(a_g), add_g(a_g), del_g(a_g) \rangle$ is defined as a 3-tuple of its grounded $pre, add$, and $del$ predicates, and we define $\mathcal{A}_g$ as the set of all grounded actions. Finally, $s_0 \subseteq \mathcal{F}_g$ is the initial state of the world for the planning task and $S_{\ast} \subseteq S$ is the set of possible goal states. 

For a grounded action $a_g \in \mathcal{A}_g$ and a state $s \in S$, we say that $a_g$ is {\em applicable} in $s$ if $pre_g(a_g) \subseteq s$. Applying an applicable action $a_g$ in the state $s$ results in a state $s[a_g] := (s / del_g(a_g)) \cup add_g(a_g)$. 
A plan for a problem $\Pi$ 
is, therefore
a sequence of grounded actions $\pi = (a_1, \cdots , a_n)$ which when applied transforms the initial state $s_0$ into a goal state in $S_{\ast}$. The action sequence defines a state sequence $\mathbf{S} = (s_0, \cdots, s_n)$ such that $s_i = s_{i-1}[a_i]$ for $1\leq i\leq n$ and $s_n\in S_{\ast}$. The set of all plans for $\Pi$ is denoted by $\mathcal{P}_{\Pi}$.

A \textit{planning domain} for a lifted STRIPS planning problem $\Pi$ is the problem's predicate and action schema sets $\mathbf{D} = 
\langle \mathcal{F}, \mathcal{A} \rangle$, while we say $\Pi$ is a problem for $\mathbf{D}$ and write $\Pi_\mathbf{D}$ if $\Pi$ uses $\mathbf{D}$ as its underlying domain, regardless of the specific objects, initial state, and goal states ($C$, $s_0$, $S_{\ast}$) for the problem.

\subsubsection{Large Language Models (LLMs)}
\textit{Language Models} are probabilistic predictors for language tokens that when given a sequence of tokens $T = (t_0, t_1, \cdots, t_n)$ in a corpus $C$ will output a set of predictions and associated probabilities $P \subseteq C \times {\rm I\!R}$ for $t_{n+1}$ based on the data the model has been trained on. Different \textit{decoding strategies} can be used to select a token in $P$ based on the probabilities, one such strategy is the greedy strategy which sets $t_{n+1}$ equal to the highest probability token in $P$. The new $t_{n+1}$ can be appended to $T$ and the process can be repeated for the next token. The maximum allowed size of $T$ is known as the \textit{context window}, which limits the amount of tokens able to use for prediction.  

\textit{Large language models} are a class of language models characterized by their large size and emergent abilities on tasks that smaller language models are unable to perform on. LLMs are almost always implemented on top of a Transformer architecture \cite{Vaswani2017}. There are many different types of large language models trained on various types of data, and models may be tuned to perform different types of tasks such as code generation \cite{Li2023StarCoderM} or acting as chat agents \cite{Touvron2023LLaMAOA}; a survey can be found at \citet{Zhao2023}.

\textit{In-Context Learning} for LLM inference is a technique classified as an emergent ability of LLMs to perform at a higher level of performance on tasks using examples of desired inputs and outputs \cite{Dong2022ASO}. For example, rather than the prompt: ``\texttt{Solve the following addition problem: 1 + 2}'', an in-context learning prompt would read: ``\texttt{Solve the following addition problems: In: 2 + 3, Out: 5; In: 4 + 2, Out: 6; ..., In: 1 + 2,}'', where the prompt is composed of 3 parts (1) An \textit{instruction} (2) a set of \textit{context examples} and a (3) a \textit{query} which is expected to be answered inline with the context examples. In-context learning is used in our work and much of the related work such as \citet{liu2023llmp} and \citet{Guan2023}.

\section{Approach}

The goal of this work is the evaluation of LLM's abilities to generate PDDL domains. In particular, we are interested in generating and evaluating these domains on an action-by-action basis where each prompt to the LLM is a request to generate one action in a domain using context examples from other domains. This action-by-action prompting was inspired by \citet{Guan2023} and is primarily a concern due to the size of the LLM's context window. 

\begin{listing}[!h]
\caption{A context example from a prompt $N(a)$ for the fly-airplane action from the logistics domain, including the "Allowed Predicates" which function as the domain specification $\langle\mathcal{F},N(\mathcal{F})\rangle$.}
\vspace{0.3cm}
\scriptsize
\begin{lstlisting}
Allowed Predicates:
(in-city ?loc - place ?city - city) : a place loc is in a city.
(at ?obj - physobj ?loc - place) : a physical object obj is at a place loc. 
(in ?pkg - package ?veh - vehicle) : a package pkg is in a vehicle veh.

Input:
The action, "FLY-AIRPLANE" will fly an airplane from one airport to another. After the action, the airplane will be in the new location. 

PDDL Action:
(:action FLY-AIRPLANE
    :parameters (?airplane - airplane ?loc-from - airport ?loc-to - airport)
    :precondition (at ?airplane ?loc-from)
    :effect (and (not (at ?airplane ?loc-from)) (at ?airplane ?loc-to))
)
\end{lstlisting}
\end{listing}

\begin{figure}[t]
\scriptsize
\centering
\includegraphics[width=0.43\textwidth]{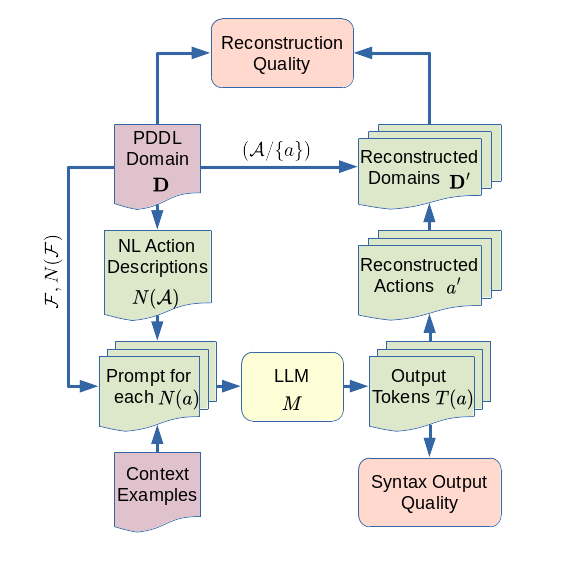}
\caption{A high-level overview of our proposed task.}
\label{fig:Arch}
\end{figure}

We now turn to characterizing the concrete task we are trying to solve, an overview can be seen in Figure \ref{fig:Arch}. In order to evaluate generated domains automatically, a ground truth domain is needed to compare the generated domains against. For this, we use existing PDDL domains as a starting point in our approach. Given a starting domain $\mathbf{D} = \langle \mathcal{F}, \mathcal{A} \rangle$, we begin by converting all action schema in $\mathcal{A}$ to natural language descriptions of action schema, $N(\mathcal{A})$. We assume that a list of the predicates in the domain $\mathcal{F}$ and natural language descriptions of these predicates $N(\mathcal{F})$ are given to us as context for the domain. This assumption, while slightly limiting accessibility, is the cornerstone that allows this task a much more robust set of automatic evaluation options than when the context for the domain is just a natural language description, as in, e.g., \citet{Guan2023}. The natural language action $N(a) \in N(\mathcal{A})$, along with a specification of domain predicates $\langle\mathcal{F},N(\mathcal{F})\rangle$, is used as the query for the in-context learning prompt. For the prompt's context examples, other actions are randomly sampled from the action schema outside of the domain $\mathbf{D}$ of the current action. A model then takes these prompts and transforms them into a sequence of tokens $T(a)$ representing $a$ as a PDDL action. An attempt is made to parse $T(a)$ as a PDDL action $a'$. This is the first location at which automated evaluation is possible, as there are numerous reasons why $T(a)$ may fail to be a valid PDDL action, many of which can be extracted by just attempting to parse $T(a)$. For all $T(a)$ that were successfully parsed into a reconstructed PDDL action $a'$, we add them to the set of successfully reconstructed actions $\mathcal{A}'$. Next, for each $a' \in \mathcal{A}'$ we create a reconstructed domain $\mathbf{D}'$ from $\mathbf{D}$ by replacing $\mathcal{A}$ with $(A / {a})\cup a'$ where $a$ is the original action that generated $a'$. Note that for our formulation $\mathcal{A}'$ is not the set of actions for a $\mathbf{D}'$, rather we look at $|\mathcal{A}'|$ new domains $\mathbf{D}'$s for each action, inline with our action by action-based evaluation strategy. This is also due to practicality reasons, in order to use $\mathcal{A}'$ for $\mathbf{D}'$, all actions in the domain would need to get through the parsing phase in which $T(a)$ is converted to $a'$, this is simply not a reasonable assumption to make. Our task then, is to evaluate the quality of each $\mathbf{D}'$ with respect to $\mathbf{D}$.

\subsection{Description Classes} \label{DescriptionClasses}
We investigate several strategies for converting PDDL action schema $a \in \mathcal{A}$ to their natural language descriptions, $N(a) \in N(\mathcal{A})$. Each strategy produces a distinct class of natural language representations of the action model.
\begin{enumerate}
    \item \textbf{Base} $N_{\text{b}}(\mathcal{A})$:
    Base descriptions include only information including the action name, parameters, and the parameter types of the action, as well as a one-line description of what the action does without explicitly mentioning any predicates. For example: \textit{``The action 'unstack' will have a hand unstack a block x from a block y.''}
    \item \textbf{Flipped} $N_{\text{f}}(\mathcal{A})$: Flipped descriptions include the base descriptions with an additional description of all predicates that are deleted preconditions in that action schema, that is, for an action schema $a \in \mathcal{A}$, $N_{\text{f}}(a)$ is $N_{\text{b}}(a)$ extended with a description of predicates in $pre(a) \cap del(a)$ as preconditions.   
    The motivation behind this class is to evaluate if predicates that are explicitly changed are the most important things to include in a natural language description for the LLM, as they might be for a person when describing a domain. For example: \textit{``The action 'unstack' will have a hand unstack a block x from a block y, if the block x is clear, on top of y, and the hand is empty.''}
    \item \textbf{Random} $N_\text{r}(\mathcal{A})$: Random descriptions act as a random baseline to compare against flipped descriptions, as well as another higher information content baseline to compare against base descriptions. For each action schema $a$, the description includes the base description $N_{\text{b}}(a)$, and descriptions of $|pre(a) \cap del(a)|$ random predicates sampled from $pre(a), add(a)$ and $del(a)$, where is the description is dependent on if the predicate was sampled from the precondition or effect.  For example: \textit{``The action 'unstack' will have a hand unstack a block x from a block y, if the hand is empty and x is on y. After the action, y should be clear.''}.
\end{enumerate}

\subsection{Evaluating} 
When considering how to evaluate the performance of LLMs on this task, 
note that
LLMs will frequently output sequences of tokens for our evaluation that cannot be interpreted as a valid planning domain. Some of these errors are syntax based while others are based on the semantics of the underlying PDDL tokens. If a model does output a valid domain, it must be evaluated in terms of its quality. 

\subsection{Domain Reconstruction Quality Metrics}

Evaluating the quality of a correctly generated planning domain is a difficult task. Current metrics such as human expert evaluation \cite{Guan2023, Li2023, Hayton2020} provide a rough but subjective measure that is impossible to automate. Like \citet{Guan2023}, we have designed our task such that all generated domains are based on an existing domain which we can evaluate with respect to a baseline. We look at and evaluate two automated metrics for measuring the quality of generated domains. The first metric, action reconstruction error, is a more traditional automated metric that measures the distance between generated actions in domains, but we note it is a poor metric. We propose a second metric, heuristic domain equivalence, which provides a more robust and tolerant approximation of true domain equivalence.  

\subsubsection{Action Reconstruction Error (ARE)}

The \textit{Action Reconstruction Error} (ARE) is a measure of how different two action schema $a, a' \in \mathcal{A}$ are. We define the action reconstruction error  as the size of the difference of predicates in the precondition and effect between $a$ and $a'$:
\begin{align*}
    \mbox{ARE}(a, a') = & |pre(a) \; \triangle \; pre(a')| + \nonumber \\
     & |add(a)  \; \triangle \; add(a')| + \\
     & |del(a) \; \triangle \; del(a')| \nonumber
\end{align*}

%
where $A \triangle B$ is the symmetric difference $(A / B) \cup (B / A)$.  This metric is useful for understanding the distribution of how close is the output domain (from the model) to the original domains. However, we claim that this metric is not a good measurement of actual domain quality. It does not take into account the fact that preconditions and effects can be added or removed from an action without changing the meaning of the action at all, for example, adding a static predicate from a precondition as an effect. To remedy this, we propose an alternative metric based on how usable the domain is for planning.

\subsubsection{Plan Applicability for Heuristic Domain Equivalence}
The primary reason a planning domain is created is so that it can be used as the underlying representation for a set of problems in the domain. The problems implicitly define a set of plans, and when reconstructing domains, we can measure domain equivalence in terms of the equivalence of the sets of plans for a collection of problems. 
%
While it is not practical to check if the full set of plans is equivalent, it is possible to check for a number of plans on some problems we care about in the domain.  

The domain equivalence heuristic is computed as follows: given an original planning domain $\mathbf{D}$, a reconstructed planning domain $\mathbf{D}'$, and a set of solvable planning problems for $D$, $\mathbf{P}_D$, each problem $\Pi\in \mathbf{P}_D$ can be transformed into a problem $\Pi'\in \mathbf{P}_{D'}$ that uses $\mathbf{D}'$ as its underlying domain. 
For each such pair of problems $\Pi$ and $\Pi'$ and some corresponding subsets of their plans $P\subseteq \mathcal{P}_{\Pi}$ and $P'\subseteq \mathcal{P}_{\Pi'}$, we can cross check whether $P\subseteq\mathcal{P}_{\Pi'}$ and 
$P'\subseteq\mathcal{P}_{\Pi}$.  For each individual plan, the test can be efficiently performed using a plan validator\footnote{\url{https://github.com/KCL-Planning/VAL}}.
%
%
This heuristic, plan equivalence on $\textbf{P}$ for a subset of plans, is a necessary condition for true domain equivalence, and its negation is a sufficient condition to show true domain inequality.

\subsection{Result Classes}
We propose four 
result classes for classifying the action from an LLMs output. Each class other than the heuristically equivalent domain class has multiple sub-classes to give a better idea of the types of problems encountered. 
\begin{enumerate}
    \item \textbf{Syntax Error}: The model produced syntactically invalid PDDL. This PDDL cannot be parsed to evaluate an action reconstruction error. Subclasses (in precedence order): (1) No PDDL (NoPDDL): Model did not output any PDDL, (2) Parenthesis Mismatch (PError): issues regarding the matching parenthesis in the PDDL (3) Unexpected Token (UToken): The PDDL parser failed after finding an unexpected token.   
    \item \textbf{Semantic Error}: The model produced syntactically valid PDDL, but the PDDL doesn't integrate with the intended problems. Subclasses (1) Type Error (TError): The model produced an unexpected type (2) Predicate Argument Error (PAError): the wrong number of variables were passed to a predicate (3) Wrong Action Name (NError), The name of the action is wrong (4) Bad Precondition (BPError): PDDL STRIPS does not allow negated preconditions, but one is present.
    \item \textbf{Different Domain}: The model produced syntactically valid PDDL that integrates with the original domain, but the underlying domains are different by way of the domain equivalence heuristic. The behavior of the actions is not as intended, plans from the original domain cannot be applied in the new domain and vice versa.
    Subclasses (1) No Plans Found (NoPlan): No plans were able to be found on problems in the new domain (2) New Plan Application Error (NPApp): Could not apply a new plan to the original domain (3) Original Plan Application Error (OPApp): The original plan could not be applied to the new domain.
    \item \textbf{(Heuristically) Equivalent Domain}: The model produced syntactically valid PDDL that integrates with the desired domain under the domain equivalence heuristic, plans from the original domain can be applied in the new domain and vice versa.
\end{enumerate}
The classes form a hierarchy in which syntax errors supersede semantic errors which supersede both the different and equivalent domain classes which are mutually exclusive. \textit{i.e.} An output with both syntax and semantic errors will only be marked with the error caught first, the syntax error.

\section{Experiments and Results}

\subsection{Setup}
We evaluate the LLaMA family of LLMs \cite{Touvron2023LLaMAOA}, as well as StarCoder (SC) \cite{Li2023StarCoderM}. For LLaMA we evaluate both the base pre-trained models at 7b, 13b, 70b parameters. We also evaluate the 7b, 13b, 70b LLaMA models that have been finetuned for chat using reinforcement learning with human feedback (RLHF) \cite{Ouyang2022TrainingLM}. For token selection for all models, we use greedy sampling in which the token with the highest output probability is selected as the next token. 

For our domains, we select 9 PDDL domains with varying action and predicate complexities. We include 2 recent domains, ``Forest" and ``Delivery" from \citet{Yang2022}, a domain ``Heavy" from \citet{Silver2023GeneralizedPI} and a novel domain, ``Trackbuilding". The latter two domains are guaranteed not to be in the training set, as they were created after LLaMA and StarCoder were trained; these domains are marked with a dagger (\dag). The remainder of our domains are famous classical planning domains from various  International Planning Competitions.

\begin{enumerate}
    \item Blocksworld – 5 predicates 4 actions:             A robot hand tries to stack blocks on a table in a particular configuration.
    \item Gripper – 4 predicates 3 actions:           A robot moves balls from one room to another using grippers.
    \item Heavy\dag – 5 predicates 2 actions:   Specified items must be packed into a box depending on item weight. 
    \item Forest – 5 predicates 2 actions: Hikers must navigate to a location over varying terrain. 
    \item Logistics – 3 predicates 6 actions: Items must be transported to locations using planes and trucks.
    \item Depot – 6 predicates 5 actions: A combination of blocks and logistics domains.
    \item Miconic – 6 predicates 4 actions: A lift delivers multiple passengers to their desired floors from their starting floors.
    \item Trackbuilding\dag – 4 predicates 3 actions: An agent must build a path for a train to take to a given location.
    \item Delivery - 7 predicates 3 actions: A delivery person must deliver newspapers to a number of safe locations from a home base.
\end{enumerate}
For the domain equivalence heuristic, our problem set consists of 2 simple randomly selected problems from each domain. We select the top $100$ plans using the K$^\ast$ planner~\cite{lee-et-al-socs2023}. The top-k plans for a problem $\Pi$ are the set of $k$ different plans with the lowest costs, which in our case is the same as the length of the plan. While any $k$ plans could be used for computing the domain equivalence heuristic, using the top-k plans
we ensure that minimally the optimal plans for the evaluated problems are equivalent. To test for plan validity we use VAL.


\subsection{Evaluating Heuristic Domain Equivalence Over Different LLMs}

\begin{figure*}[t]
\centering
\includegraphics[width=1\textwidth]{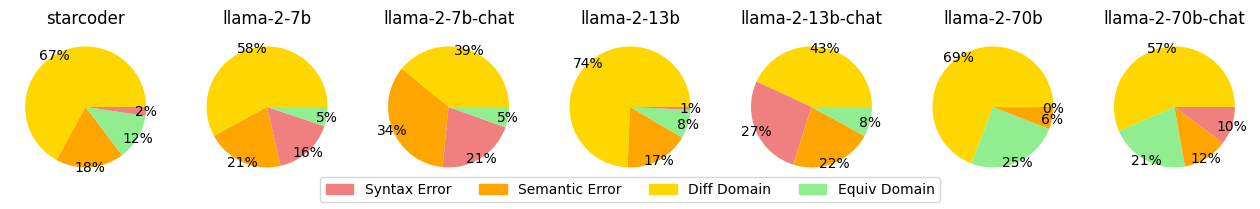}
\includegraphics[width=1\textwidth]{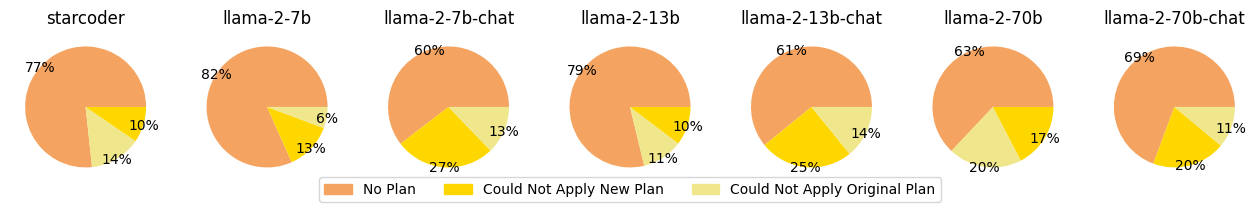}
\caption{(Top) Characterizing LMM outputs in terms of core result classes. (Bottom) Breakdown of Diff domain subclasses.}
\label{fig:Main}
\end{figure*}

\begin{figure}[t]
\centering
\includegraphics[width=0.4\textwidth]{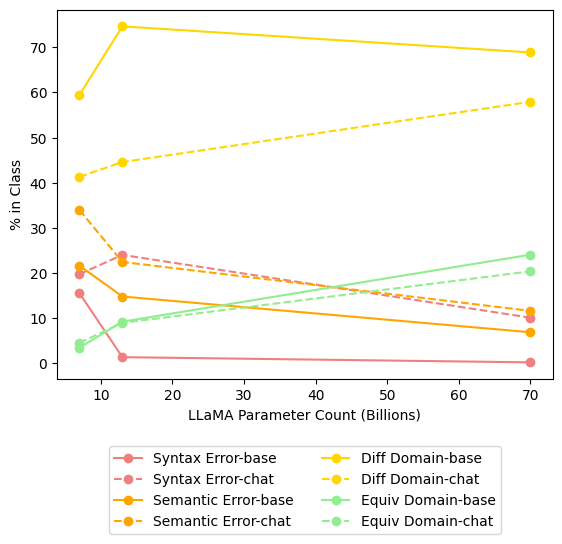}
\caption{Overview of LLaMA result class percentages with respect to model size. Contains both chat and base models.}
\label{fig:LLaMAParams}
\end{figure}

For this experiment, we exclusively use base descriptions in which only a description of the action's parameters and types without reference to predicates is provided. For prompt generation, each base action description is turned into 60 prompts, each with 3 randomly sampled context examples from outside of its domain. We note that this sampling is done uniformly across all types of actions, the only restriction being that the action used for context cannot be in the same domain as the action we are generating for. We chose to use 60 prompts as a trade-off between experiment runtime and statistical significance. We chose to use 3 context examples after a manual parameter search; increasing the number of context examples further did not improve results and decreasing past 3 led to worse results.    

Figure \ref{fig:Main} (Top) displays the breakdown of outputs over the primary result classes. Two results are immediately apparent from this. First, LLMs particularly larger ones, are quite good at generating syntactically and semantically valid PDDL, the best model LLaMA-2-70b, is able to construct valid PDDL in 94\% of domains. When looking at valid PDDL generated, we see that the ratio of heuristically equivalent domains to non-equivalent domains and the number of heuristically equivalent domains is largely dependent on model size (see Figure \ref{fig:LLaMAParams}). The best result was on LLaMA-2-70b. It reconstructed 25\% domains to be heuristically equivalent to the natural language descriptions. This is a very promising result in terms of the applicability of LLMs for the task of PDDL domain generation. Second, in terms of different types of models, it is surprising that the LLaMA chat models perform worse on this task than base LLaMA models across the board. Typically these models that have been trained with RLHF are seen to do better than base models across the board \cite{Ouyang2022TrainingLM}.

\begin{table}[!t]
   \def\arraystretch{1.32}
\setlength\tabcolsep{3pt}
    \centering
    \resizebox{\columnwidth}{!}{
    \begin{tabular}{l|l||l|l|l|l|l|l|l}
   \multicolumn{2}{c||}{  Result class} & Star &\multicolumn{6}{c}{  LLaMA} \\
       \multicolumn{2}{c||}{  \& subclass} & Coder & 7b & 7b-C & 13b & 13b-C & 70b & 70b-C  \\ \hline
        \hline
        \multirow{4}{*}{\rotatebox{90}{Syntax}}  
        & NoPDDL & \textbf{0.00} & 0.16 & \textbf{0.00} & \textbf{0.00}  & \textbf{0.00}  & \textbf{0.00}  & 0.16 \\
        & PError & \textbf{0.00} & \textbf{0.00} & 0.78 & \textbf{0.00}  & \textbf{0.00}  & \textbf{0.00}  & \textbf{0.00} \\
        & UToken & 2.34 & 16.09 & 20.31 & 0.78  & 27.03  & \textbf{0.31}  & 10.00 \\ 
        \cline{2-9}
        & Total & 2.34 & 16.25 & 21.09 & 0.78  & 27.03  & \textbf{0.31}  & 10.16 \\ 
        \hline
        \hline
        \multirow{5}{*}{\rotatebox{90}{Semantics}}   &PAError & 16.56 & 16.56 & 21.56 & 11.41  & 15.16  & \textbf{4.22}  & 10.78 \\
        &NError & \textbf{0.00} & 0.47 & 0.62 & \textbf{0.00}  & 0.62  & \textbf{0.00}  & \textbf{0.00} \\
        &TError & 1.72 & 3.59 & 12.03 & 5.62  & 6.09  & 1.41  & \textbf{1.25} \\
        &BPError & \textbf{0.00} & \textbf{0.00} & 0.16 & 0.16  & \textbf{0.00}  & \textbf{0.00}  & \textbf{0.00} \\ 
        \cline{2-9}
        & Total & 18.28 & 20.62 & 34.38 & 17.19  & 21.88  & \textbf{5.62}  & 12.03 \\ 
        \hline
        \hline
        
        \multirow{4}{*}{\rotatebox{90}{Diff}} &NoPlan & 51.41 & 47.34 & \textbf{23.59} & 58.59  & 26.25  & 43.59  & 39.22 \\
        &NPApp & \textbf{6.41} & 7.34 & 10.47 & 7.66  & 10.78  & 12.03  & 11.09 \\
        &OPApp & 9.22 & \textbf{3.28} & 5.00 & 8.12  & 6.09  & 13.59  & 6.25 \\ 
        \cline{2-9}
        & Total & 67.03 & 57.97 & \textbf{39.06} & 74.38  & 43.12  & 69.22  & 56.56 \\ \hline
        \hline
        \multicolumn{2}{c||}{Equiv} & 12.34 & 5.16 & 5.47 & 7.66  & 7.97  & \textbf{24.84}  & 21.25 \\
    \end{tabular}
    }
    \caption{Distribution of result classes and subclasses. Lower is better
    for all classes and subclasses except equivalent domain (Equiv), for which higher is better. Best results in bold.
    }
    \label{table:syntaxsemErrors}
\end{table}

We next turn to discuss result subclasses. Table \ref{table:syntaxsemErrors} displays the lopsided breakdown of syntax and semantic errors. There were almost no instances of the No PDDL subclass, all models evaluated output something minimally interpretative as PDDL; except LLaMA 7b. Parenthesis mismatch errors (PError) were also negligible. The overwhelming majority of syntax errors were unexpected token errors (UToken). This encompassed a whole range of issues from duplicate ``\texttt{:precondition}'' tags to attempting to add type annotation to variables mentioned in predicates. For semantic errors, the primary breakdown was dominated by issues related to predicate-argument (PAError) counts where the model added or removed arguments to predicates in the action schema. Type errors (TError) were rare, we note that LLAMA 70b Chat performed best in this regard. Incorrect action name errors (NError) were exceedingly rare and Bad Precondition (BPError) was rarer still. 
Of the semantically and syntactically valid domains, the majority were different domains.
Different-domain subclasses displayed in Figure \ref{fig:Main} (Bottom) and Table ~\ref{table:syntaxsemErrors} reveal an interesting insight into the quality of generated domains. The results show that across the board, the different domains could not be used for planning; the planner failed to produce any valid plan using the reconstructed problems in the domain $P_{D'}$ (NoPlan).
%
The remaining different-domains failures are split relatively equally due to failures in cross-validating the new plans on the original domains (NPApp) and vice versa (OPApp).     

\subsection{Evaluating Heuristic Domain Equivalence Over Description Classes and LLMs}

\begin{figure*}[t]
\centering
\includegraphics[width=1\textwidth]{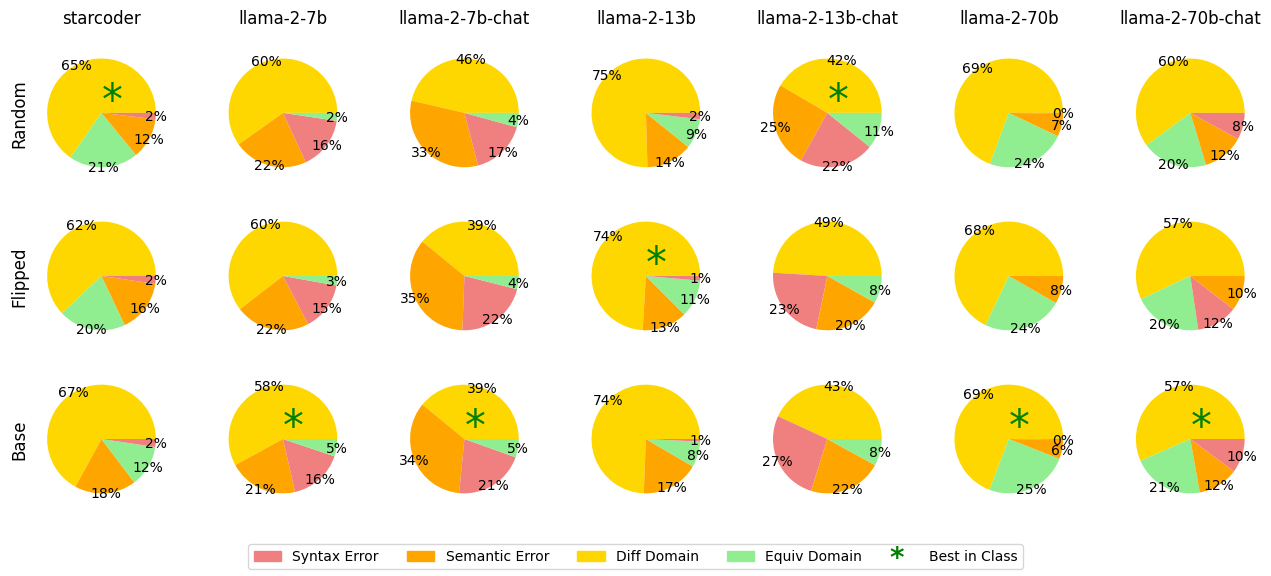}
\caption{Breakdown of LLMs over top level result classes vs different description classes.}
\label{fig:DescriptionClass}
\end{figure*}

For this experiment, we evaluate result classes over the three proposed description classes. To generate our prompts, we map each action to 20 prompts in each of the 3 description classes. The context for the prompts is taken from the same description class and is always taken from domains outside the domain of the action to evaluate. For evaluation, we use the same setup as our first experiment and evaluate over our result classes. Figure \ref{fig:DescriptionClass} displays a breakdown of the performance of each model on each description class. The results show that while on some models the flipped class performs well, it is not consistent and not as statistically significant as we had predicted. We are surprised to see that the base class performs on par with the random and flipped classes on the LLaMA models, leading us to conclude that at least for the classes we looked at where the number of predicates in flipped is small, the extra information provided by the random and flipped descriptions is not significant enough to sway the results for these models. The anomaly here is StarCoder in which providing the extra context in the random and flipped classes boots its performance by around 10\%. 

\subsection{Action Reconstruction Error and Result Class}
For this experiment, we evaluate the models in terms of their action reconstruction error to see how close from a predicate-by-predicate point of view the model gets to reconstructing the original actions. Additionally, we investigate the the relationship between the action reconstruction error and the result classes as well as how the action reconstruction error may be used to augment our use of heuristic domain equivalence. This experiment uses the same setup as the experiment over description classes, each $a \in \mathcal{A}$ is mapped to $N_b(a)$ and is used for 60 prompts. All prompts are evaluated on each LLM and result classes and ARE is evaluated for classes for all classes except syntax errors as ARE cannot be automatically computed without a parsed action.   

\begin{figure*}[t]
\centering
\includegraphics[width=1\textwidth]{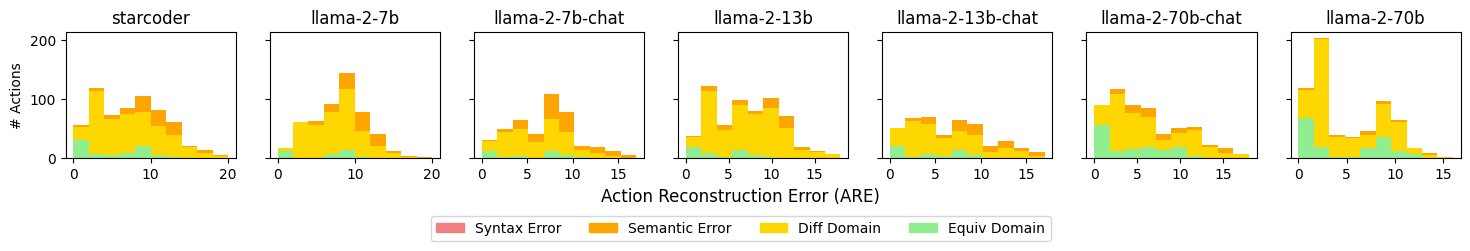}
\caption{Action Reconstruction Error (ARE) distribution with respect to reconstruction class over LLMs.}
\label{fig:ARE}
\end{figure*}

Figure \ref{fig:ARE} displays the distributions of action reconstruction errors (ARE) for each model, and splits each bucket by reconstruction class. This gives a good picture of how much each model deviates from the original action. We note that the better-performing models tend to have their distributions cluster around lower AREs, that is, they construct actions that are similar in terms of the exact predicates used in the original action. This additionally exposes the flaws of ARE as a metric for domain equivalence as we can see that just being close to the original action in terms of predicate similarity is not good enough and that plenty of domains outside this range are heuristically equivalent. This understanding of ARE can also help us find false positives in heuristically equivalent domains that are not truly equal since only a finite number of problems and plans for each problem can be evaluated. Hence when searching for false positives it can be useful to start with domains with the highest ARE since it is more likely something with many predicates changed from the original action represents a different domain.

\section{Related Work}

\subsection{Large Language Models and Planning}
There are been a number of papers that investigate the use of LLMs for planning. Some recent work (cf. \citet{Valmeekam2022LargeLM, Raman2022}) use LLMs as planners, while others (cf \citet{Guan2023, liu2023llmp}) use LLMs as auxiliary components of a hybrid planning system while leveraging automated planners for solving the planning task. The general consensus seems to be that LLMs are not very good as planners. This finding was one of the motivations for this work in this work, as we focus on using LLMs to aid automated planning rather than as planners themselves.  

\subsubsection{LLM+P} 
The LLM+P framework \cite{liu2023llmp} was one of the first to recognize the potential of combining LLMs and planners as hybrid systems, and utilizing LLMs to 
east the use of automated planners
. The LLM+P architecture takes in (1) natural language descriptions of problem in a planning domain, (2) a context example of a natural language problem in the given domain being converted to a PDDL problem, and (3) a PDDL domain file. Using these inputs the model uses an underlying LLM to convert the natural language problem description and context into a PDDL problem. This is then combine with the PDDL domain input to an automated planner producing a PDDL plan, the resulting plan is then fed into an LLM which describes the plan in natural language. LLM+P's applicability is somewhat hindered by their assumptions that a PDDL domain exists, and context examples converting natural language descriptions of problems to PDDL problems for these domains exist. Such assumptions are impossible to meet in the case of things like narrative action model acquisition, and indeed still requires an expert in the system somewhere to write the domains and the context examples. Our work does not focus on using LLMs to generate PDDL tasks, but it is tangential to all of LLM+P's assumptions. We (1) investigate the construction the PDDL domain rather than have it provided and (2) do this using context examples from arbitrary domains rather than from the same domain.

\subsubsection{LLM-DM}
The most closely related work to ours is the end-to-end domain construction and planning framework from \citet{Guan2023} which we will call LLM-DM. LLM-DM is composed of a three-part process, automated domain construction, human refinement of domain, and planning with the domain. We are interested primarily in their automated domain construction as it is a very similar task to ours. For this, LLM-DM generates a domain on an action-by-action basis, each prompt containing five parts: (1) an instruction describing the PDDL creation task, (2) one or two context examples from the blocksworld domain on what a translation of an action description to PDDL looks like, (3) a natural language description of the domain, (4) a natural language description of the action and (5) a dynamically updated list of predicates used by the domain including natural language action descriptions. As the domain is generated action-by-action, the instruction and context examples include requests for the model to generate a list of new predicates based on the description of the action. LLM-DM evaluates constructing PDDL on three domains (Logistics, Tyreworld, and a custom domain, "Household") using the LLMs GPT-4 \cite{OpenAI2023GPT4TR} and GPT-3.5 Turbo (ChatGPT). To measure the quality of the constructed domain, manual human evaluation is used, experts annotate the PDDL domain output, marking the PDDL with mistakes and corrections, which the authors claim provides and approximate distance between the generated PDDL and correct PDDL. 

LLM-DM provided inspiration in our work to generate domains using LLMs on an action-by-action basis rather than trying to have the LLM output the full domain. The authors cite well-founded concerns about the context window size and the potential for corrective feedback on an action-by-action basis, making this more useful for the end user. For our work, instead of providing the model with a description of the domain and having the model extract the predicates at each stage on-top of the action translation, we explicitly provided the allowed predicates and their description \textit{as} the description of the domain. This change
is key for being able 
to automatically evaluate the constructed domains, and is responsible for our automated evaluation approaches rather than a manual evaluation approaches. 

\subsection{Textual and Narrative Action Model Acquisition}
The task we propose is similar to the action-model extraction from text task \cite{Lindsay2017} and narrative action-model acquisition task from text task \cite{Hayton2020,Li2023} in which the goal is from natural language to generate the entire domain model from $\mathcal{F}_g$ and $\mathcal{A}_g$ if grounded and $\mathcal{F}, \mathcal{A}$, and potentially $\mathcal{C}$ if lifted. A downside of these tasks is that it very difficult to automatically evaluate performance on, as it requires a full understanding of the natural language text and expert knowledge of PDDL domains. Evaluation for these tasks is frequently done either via expert analysis of the generated PDDL domain such as in \cite{Hayton2020,Huang2014} or automated metrics such as  that can't fully capture the performance of the model. These shortcomings in evaluation were a driver of both our problem formulation and proposed domain quality metrics.

\section{Conclusion and Future Work}

There are many avenues that could be explored using this work as a springboard. In particular, we are interested in three main directions: (1) deeper investigations of the capabilities of large language models in terms of selection and tuning, (2) using re-prompting for fixing mistakes in PDDL for chat-based LLMs, (3) investigating more robust tasks and metrics. 

First, in terms of LLMs there is a lot that could be done to extend this work. The results showing improved performance on larger models are a good starting point for future work and are in line with \citet{Guan2023} which evaluates with respect to GPT-4 and GPT 3.5. coming to similar conclusions that larger pre-trained models are better when it comes to handling PDDL construction. Future work and applications not interested in tuning should take this into consideration using larger models such as GPT-4 and LLaMA-70b as baselines, other large models such as Bloom \cite{bigscience_workshop_2022} would be promising to evaluate over. Our experiment over description classes revealed the coding model StarCoder performs quite well in certain cases when additional predicate information is included in natural language descriptions, we believe this warrants a further investigation of coding models and their capabilities. Beyond just the selection of LLMs, there are two more properties of LLMs we could investigate. First, LLM tuning approaches, such as finetuning and prompt tuning have been shown to allow small LLMs to perform well on tasks they are tuned on. Second, chat-based LLMs with large context windows can be re-prompt and provide corrective feedback \cite{Raman2022}. \citet{Guan2023} successfully demonstrate corrective reprompting from tools like VAL and other reprompting to provide corrective feedback to LLMs. Using our result classification system, adding support for corrective reprompting where the re-prompt is based on information regarding the result class is a clear next step. 

Finally, we discuss potential alternatives that could be made to our evaluation. As discussed in our approach, we do not use $\mathcal{A'}$ as the set of action schema for a $\mathbf{D}'$ for a number of practical reasons. However, evaluating the performance of domains in which all actions are generated is a desirable target for evaluation. Towards this end, it would be interesting to evaluate with respect to a form of iterative domain completion task after an initial action has been generated. Previously generated actions in $\mathcal{A'}$ could then be used as part of the prompt until a full reconstructed action schema for the reconstructed domain $\mathbf{D}'$ has been constructed.  

\section*{Acknowledgement}
This work is supported by IBM Research through the Rensselaer IBM AI Research Collaboration (https://airc.rpi.edu/)

\fontsize{10pt}{11pt}\selectfont
\bibliography{main}

\begin{thebibliography}{27}
\providecommand{\natexlab}[1]{#1}

\bibitem[{Biere et~al.(2021)Biere, Heule, van Maaren, and Walsh}]{SATBook}
Biere, A.; Heule, M.; van Maaren, H.; and Walsh, T., eds. 2021.
\newblock \emph{Handbook of Satisfiability - Second Edition}, volume 336 of
  \emph{Frontiers in Artificial Intelligence and Applications}.
\newblock {IOS} Press.

\bibitem[{{BigScience Workshop}(2022)}]{bigscience_workshop_2022}
{BigScience Workshop}. 2022.
\newblock {BLOOM} (Revision 4ab0472).

\bibitem[{Corr{\^{e}}a and Seipp(2022)}]{Corra2022BestFirstWS}
Corr{\^{e}}a, A.~B.; and Seipp, J. 2022.
\newblock Best-First Width Search for Lifted Classical Planning.
\newblock In \emph{{ICAPS}}, 11--15. {AAAI} Press.

\bibitem[{Dong et~al.(2022)Dong, Li, Dai, Zheng, Wu, Chang, Sun, Xu, and
  Sui}]{Dong2022ASO}
Dong, Q.; Li, L.; Dai, D.; Zheng, C.; Wu, Z.; Chang, B.; Sun, X.; Xu, J.; and
  Sui, Z. 2022.
\newblock A Survey on In-context Learning.

\bibitem[{Ghallab et~al.(1998)Ghallab, Howe, Knoblock, McDermott, Ram, Veloso,
  Weld, and Wilkins}]{mcdermott-et-al-tr1998}
Ghallab, M.; Howe, A.; Knoblock, C.; McDermott, D.; Ram, A.; Veloso, M.; Weld,
  D.; and Wilkins, D. 1998.
\newblock PDDL --- The Planning Domain Definition Language.
\newblock \emph{Technical Report}.

\bibitem[{Ghallab, Nau, and Traverso(2004)}]{GhallabNT}
Ghallab, M.; Nau, D.~S.; and Traverso, P. 2004.
\newblock \emph{Automated planning - theory and practice}.
\newblock Elsevier.

\bibitem[{Guan et~al.(2023)Guan, Valmeekam, Sreedharan, and
  Kambhampati}]{Guan2023}
Guan, L.; Valmeekam, K.; Sreedharan, S.; and Kambhampati, S. 2023.
\newblock Leveraging Pre-trained Large Language Models to Construct and Utilize
  World Models for Model-based Task Planning.
\newblock In \emph{NeurIPS}.

\bibitem[{Harrison, Urban, and Wiedijk(2014)}]{HarrisonUW14}
Harrison, J.; Urban, J.; and Wiedijk, F. 2014.
\newblock History of Interactive Theorem Proving.
\newblock In \emph{Computational Logic}, volume~9 of \emph{Handbook of the
  History of Logic}, 135--214. Elsevier.

\bibitem[{Hayawi, Shahriar, and Mathew(2024)}]{ImitationGames}
Hayawi, K.; Shahriar, S.; and Mathew, S.~S. 2024.
\newblock The imitation game: Detecting human and AI-generated texts in the era
  of ChatGPT and BARD.
\newblock \emph{Journal of Information Science}.

\bibitem[{Hayton et~al.(2020)Hayton, Porteous, Ferreira, and
  Lindsay}]{Hayton2020}
Hayton, T.; Porteous, J.; Ferreira, J.~F.; and Lindsay, A. 2020.
\newblock Narrative Planning Model Acquisition from Text Summaries and
  Descriptions.
\newblock In \emph{{AAAI}}, 1709--1716. {AAAI} Press.

\bibitem[{Helmert(2006)}]{Helmert06}
Helmert, M. 2006.
\newblock The Fast Downward Planning System.
\newblock \emph{J. Artif. Intell. Res.}, 26: 191--246.

\bibitem[{Huang, Chen, and Zhang(2014)}]{Huang2014}
Huang, R.; Chen, Y.; and Zhang, W. 2014.
\newblock SAS+ Planning as Satisfiability.
\newblock \emph{J. Artif. Intell. Res.}, 43: 293--328.

\bibitem[{Lee, Katz, and Sohrabi(2023)}]{lee-et-al-socs2023}
Lee, J.; Katz, M.; and Sohrabi, S. 2023.
\newblock On K* Search for Top-k Planning.
\newblock In \emph{Proceedings of the 16th Annual Symposium on Combinatorial
  Search (SoCS 2023)}. {AAAI} Press.

\bibitem[{Li et~al.(2023)Li, {Ben allal}, Zi, Muennighoff, Kocetkov
  et~al.}]{Li2023StarCoderM}
Li, R.; {Ben allal}, L.; Zi, Y.; Muennighoff, N.; Kocetkov, D.; et~al. 2023.
\newblock StarCoder: may the source be with you!
\newblock \emph{Transactions on Machine Learning Research}.
\newblock Reproducibility Certification.

\bibitem[{Li et~al.(2024)Li, Cui, Lin, and Haslum}]{Li2023}
Li, R.; Cui, L.; Lin, S.; and Haslum, P. 2024.
\newblock NaRuto: Automatically Acquiring Planning Models from Narrative Texts.
\newblock In \emph{{AAAI}}, volume~38, 20194--20202. {AAAI} Press.

\bibitem[{Lindsay et~al.(2017)Lindsay, Read, Ferreira, Hayton, Porteous, and
  Gregory}]{Lindsay2017}
Lindsay, A.; Read, J.; Ferreira, J.~F.; Hayton, T.; Porteous, J.; and Gregory,
  P. 2017.
\newblock Framer: Planning Models from Natural Language Action Descriptions.
\newblock In \emph{{ICAPS}}, 434--442. {AAAI} Press.

\bibitem[{Liu et~al.(2023)Liu, Jiang, Zhang, Liu, Zhang, Biswas, and
  Stone}]{liu2023llmp}
Liu, B.; Jiang, Y.; Zhang, X.; Liu, Q.; Zhang, S.; Biswas, J.; and Stone, P.
  2023.
\newblock LLM+P: Empowering Large Language Models with Optimal Planning
  Proficiency.
\newblock arXiv:2304.11477.

\bibitem[{OpenAI(2023)}]{OpenAI2023GPT4TR}
OpenAI. 2023.
\newblock GPT-4 Technical Report.
\newblock arXiv:2303.08774.

\bibitem[{Ouyang et~al.(2022)Ouyang, Wu, Jiang, Almeida
  et~al.}]{Ouyang2022TrainingLM}
Ouyang, L.; Wu, J.; Jiang, X.; Almeida, D.; et~al. 2022.
\newblock Training language models to follow instructions with human feedback.
\newblock In \emph{NeurIPS}.

\bibitem[{Raman et~al.(2022)Raman, Cohen, Rosen, Idrees, Paulius, and
  Tellex}]{Raman2022}
Raman, S.~S.; Cohen, V.; Rosen, E.; Idrees, I.; Paulius, D.; and Tellex, S.
  2022.
\newblock Planning With Large Language Models Via Corrective Re-Prompting.
\newblock In \emph{NeurIPS 2022 Foundation Models for Decision Making
  Workshop}.

\bibitem[{Silver et~al.(2024)Silver, Dan, Srinivas, Tenenbaum, Kaelbling, and
  Katz}]{Silver2023GeneralizedPI}
Silver, T.; Dan, S.; Srinivas, K.; Tenenbaum, J.~B.; Kaelbling, L.~P.; and
  Katz, M. 2024.
\newblock Generalized Planning in PDDL Domains with Pretrained Large Language
  Models.
\newblock In \emph{AAAI Conference on Artificial Intelligence (AAAI)}. {AAAI}
  Press.

\bibitem[{Touvron, Lavril, and Izacard(2023)}]{Touvron2023LLaMAOA}
Touvron, H.; Lavril, T.; and Izacard, G. 2023.
\newblock LLaMA: Open and Efficient Foundation Language Models.
\newblock arXiv:2302.13971.

\bibitem[{Valmeekam et~al.(2023)Valmeekam, Marquez, Sreedharan, and
  Kambhampati}]{Valmeekam2022LargeLM}
Valmeekam, K.; Marquez, M.; Sreedharan, S.; and Kambhampati, S. 2023.
\newblock On the Planning Abilities of Large Language Models - A Critical
  Investigation.
\newblock In Oh, A.; Neumann, T.; Globerson, A.; Saenko, K.; Hardt, M.; and
  Levine, S., eds., \emph{Advances in Neural Information Processing Systems},
  volume~36, 75993--76005. Curran Associates, Inc.

\bibitem[{Vaswani et~al.(2017)Vaswani, Shazeer, Parmar, Uszkoreit, Jones,
  Gomez, Kaiser, and Polosukhin}]{Vaswani2017}
Vaswani, A.; Shazeer, N.~M.; Parmar, N.; Uszkoreit, J.; Jones, L.; Gomez,
  A.~N.; Kaiser, L.; and Polosukhin, I. 2017.
\newblock Attention is All you Need.
\newblock In \emph{NIPS}.

\bibitem[{Yang et~al.(2022)Yang, Silver, Curtis, Lozano{-}P{\'{e}}rez, and
  Kaelbling}]{Yang2022}
Yang, R.; Silver, T.; Curtis, A.; Lozano{-}P{\'{e}}rez, T.; and Kaelbling,
  L.~P. 2022.
\newblock {PG3:} Policy-Guided Planning for Generalized Policy Generation.
\newblock In \emph{{IJCAI}}, 4686--4692.

\bibitem[{Zhao et~al.(2023{\natexlab{a}})Zhao, Zhou, Li, Tang, Wang, Hou, Min,
  Zhang, Zhang, Dong, Du, Yang, Chen, Chen, Jiang, Ren, Li, Tang, Liu, Liu,
  Nie, and rong Wen}]{Zhao2023}
Zhao, W.~X.; Zhou, K.; Li, J.; Tang, T.; Wang, X.; Hou, Y.; Min, Y.; Zhang, B.;
  Zhang, J.; Dong, Z.; Du, Y.; Yang, C.; Chen, Y.; Chen, Z.; Jiang, J.; Ren,
  R.; Li, Y.; Tang, X.; Liu, Z.; Liu, P.; Nie, J.; and rong Wen, J.
  2023{\natexlab{a}}.
\newblock A Survey of Large Language Models.
\newblock arXiv:2303.18223.

\bibitem[{Zhao et~al.(2023{\natexlab{b}})Zhao, Song, Duah, Macbeth, Carter,
  Van, Bravo, Klenk, Sick, and Filipowicz}]{ZhaoSDMCVBKSF23}
Zhao, Z.; Song, S.; Duah, B.; Macbeth, J.; Carter, S.~A.; Van, M.~P.; Bravo,
  N.~S.; Klenk, M.; Sick, K.; and Filipowicz, A. L.~S. 2023{\natexlab{b}}.
\newblock More human than human: LLM-generated narratives outperform human-LLM
  interleaved narratives.
\newblock In \emph{Creativity {\&} Cognition}, 368--370. {ACM}.

\end{thebibliography}

\appendix

\section{Sample Prompt}
The prompt is composed of three parts:
\begin{enumerate}
    \item \textbf{Instruction}: A description of the task.
    \item \textbf{Context Examples (x3)}: Example problems with inputs and correct outputs, specifies the problem format. Each of our context examples contains three parts: (1) The list of allowed predicates and their natural language descriptions in the domain (we refer to this as $\langle\mathcal{F},N(\mathcal{F})\rangle$ in the paper). (2) The natural language description of the action, varied according to description class. (3) The PPDL of the action. 
    \item \textbf{Query}: The input for the action we want to generate the PDDL for. Includes both the allowed predicates and natural language input. 
\end{enumerate}
The following example prompt is part of the random description class in which a random predicate within the PDDL reference action is described. The final prompt is the concatenation of all three parts.

\begin{listing}[h!]
\caption{The Instruction. This specific instruction is used in all prompts and was selected after manually testing other similar instruction prompts with no significant changes in results.}
\scriptsize
\begin{lstlisting}
Given a description of an action in some domain, convert it to Planning Domain Definition Language (PDDL) action. You may only use the allowed predicates provided for each action.

\end{lstlisting}
\end{listing}

\begin{listing}[!h]%
\caption{Context Example 1. The fly-airplane action from the logistics domain.}
\scriptsize
\begin{lstlisting}
Allowed Predicates:
(in-city ?loc - place ?city - city) : a place loc in in a city.
(at ?obj - physobj ?loc - place) : a physical object obj is at a place loc. 
(in ?pkg - package ?veh - vehicle) : a package pkg is in a vehicle veh.

Input:
The action, "FLY-AIRPLANE" will fly an airplane from one airport to another . After the action, the airplane will be in the new location. 

PDDL Action:
(:action FLY-AIRPLANE
    :parameters (?airplane - airplane ?loc-from - airport ?loc-to - airport)
    :precondition (at ?airplane ?loc-from)
    :effect (and (not (at ?airplane ?loc-from)) (at ?airplane ?loc-to))
)
\end{lstlisting}
\end{listing}

\begin{listing}[!h]%
\caption{Context Example 2. The pack-first action from the heavy pack domain.}
\scriptsize
\begin{lstlisting}
Allowed Predicates:
(heavier ?item1 - item ?item2 - item) : item1 is heavier than item2.
(packed ?i - item) : item i is packed into the box.
(unpacked ?i - item) : item i is unpacked from the box.
(nothing-above ?i - item) : nothing is above item i in the box.
(box-empty) : the box is empty.

Input:
The action, "pack-first" will pack an item into the box . After the action, the item will be packed.

PDDL Action:
(:action pack-first
    :parameters (?item - item)
    :precondition (box-empty)
    :effect (and (not (box-empty)) (packed ?item) (nothing-above ?item) (not (unpacked ?item)))
)
\end{lstlisting}
\end{listing}

\begin{listing}[!h]%
\caption{Context Example 3. The Load action from the heavy pack domain.}
\scriptsize
\begin{lstlisting}
Allowed Predicates:
(at ?x - locatable ?y - place) : the locatable x is at some place.
(on ?x - crate ?y - surface) : crate x is on a surface y.
(in ?x - crate ?y - truck) : crate x is in truck y. 
(lifting ?x - hoist ?y - crate) : hoist x is lifting crate y.
(available ?x - hoist) : hoist x is available.
(clear ?x - surface) : surface x is clear.

Input:
The action, "Load" will use a hoist to load a crate into a truck at a place if the truck is at the place.

PDDL Action:
(:action Load
    :parameters (?x - hoist ?y - crate ?z - truck ?p - place)
    :precondition (and (at ?x ?p) (at ?z ?p) (lifting ?x ?y))
    :effect (and (not (lifting ?x ?y)) (in ?y ?z) (available ?x))
)
\end{lstlisting}
\end{listing}

\begin{listing}[!h]%
\caption{The Query}
\scriptsize
\begin{lstlisting}
Allowed Predicates:
(handempty) : the hand is empty.
(holding ?x - block): block x is held.
(clear ?x - block): block x is clear.
(on ?x - block ?y - block) : block x is on block y.
(ontable ?x - block): block x is on the table.

Input:
The action, "put-down" will have the hand put down a block if the block is being held.

PDDL Action:

\end{lstlisting}
\end{listing}

We note that: (1) All the context examples in this sample are from different domains, but they do not have to be. The actions used for context examples are uniformly sampled from all domains except the domain being used for the query. (2) The average prompt length for all description classes was approximately 400 words, well under the context windows of all models experimented with. 

\section{Sample Responses}
Responses are the token sequence generated by the large language model in response to our prompt.  We apply simple parenthesis matching based pruning to the model's response and then try and parse it as PDDL. The pruner attempts to match the outermost parenthesis in the response and takes this sub-string of the response to parse as PDDL. Errors from this process result in a syntax error class, specifically either the NoPDDL or PError sub-classes, depending if no PDDL is found or the pruner is unable to match parenthesis respectively. The following seven responses are from the sample prompt provided above:

\begin{listing}[!h]%
\caption{Pruned StarCoder Response}
\scriptsize
\begin{lstlisting}
(:action put-down
    :parameters (?x - block)
    :precondition (and (holding ?x))
    :effect (and (handempty) (clear ?x) (not (holding ?x)))
)
\end{lstlisting}
\end{listing}

\begin{listing}[!h]%
\caption{Pruned LLaMA 7b Response}
\scriptsize
\begin{lstlisting}
(:action put-down
    :parameters (?x - block)
    :precondition (and (holding ?x) (handempty))
    :effect (and (clear ?x) (handempty))
)
\end{lstlisting}
\end{listing}

\begin{listing}[!h]%
\caption{Pruned LLaMA 7b Chat Response}
\scriptsize
\begin{lstlisting}
(:action put-down
    :parameters (?block - block)
    :precondition (handempty)
    :effect (and (not (holding ?block)) (clear ?block) (ontable ?block))
)
\end{lstlisting}
\end{listing}

\begin{listing}[!h]%
\caption{Pruned LLaMA 13b Chat Response}
\scriptsize
\begin{lstlisting}
(:action put-down
    :parameters (?x - block)
    :precondition (and (holding ?x))
    :effect (and (not (holding ?x)))
)
\end{lstlisting}
\end{listing}

\begin{listing}[!h]%
\caption{Pruned LLaMA 13b Chat Response}
\scriptsize
\begin{lstlisting}
(:action put-down
    :parameters (?x - block)
    :precondition (holding ?x)
    :effect (and (not (holding ?x)) (clear ?x))
)
\end{lstlisting}
\end{listing}

\begin{listing}[!h]%
\caption{Pruned LLaMA 70b Response}
\scriptsize
\begin{lstlisting}
(:action put-down
    :parameters (?x - block)
    :precondition (holding ?x)
    :effect (and (not (holding ?x)) (clear ?x))
)
\end{lstlisting}
\end{listing}

\begin{listing}[!h]%
\caption{Pruned LLaMA 70b Chat Response}
\scriptsize
\begin{lstlisting}
(:action put-down
    :parameters (?x - block)
    :precondition (holding ?x)
    :effect (and (not (holding ?x)) (on ?x ?table))
)
\end{lstlisting}
\end{listing}

\section{Hyperparameters}
We provide additional information on our hyperparamaters to 
assist with reproducibility. The following hyperparamters were used for all models, including StarCoder:
\begin{itemize}
    \item Decoding strategy: greedy. We select the most probable token, we found from a manual search that this produced better domains with less syntax and semantic issues. 
    \item Maximum Token Output: 300. We limit the number of new tokens created to 300. This is enough that all reasonable combinations of predicates in all domains can be captured. 
    \item Stop Sequences: "Input:" and "Allowed Predicates:", we stop generating on these sentinel tokens as they signal the PDDL response has likely been completed.
\end{itemize}


\end{document}